\documentclass{article}
\usepackage{spconf,amsmath,graphicx}
\ninept

\usepackage{cite}
\usepackage{amsmath,amssymb,amsfonts}
\usepackage{algorithmic}
\usepackage{graphicx}
\usepackage{textcomp}
\usepackage{multirow}
\usepackage{xcolor}
\usepackage{bm}
\usepackage{color}
\usepackage{multirow}
\usepackage{multicol}
\usepackage{makecell}
\usepackage{color}
\usepackage{xcolor}
\usepackage{colortbl}
\usepackage{bm}
\usepackage{amsmath}
\usepackage{amssymb}
\usepackage{graphicx}
\usepackage{extarrows}
\usepackage{tcolorbox}
\usepackage{hhline}
\usepackage{enumitem}
\definecolor{lightgray}{rgb}{0.88, 0.92, 0.98}
\definecolor{defblue}{rgb}{0.1843, 0.3333, 0.6}
\definecolor{defred}{rgb}{0.88, 0.2510, 0.3294}

\definecolor{green1}{rgb}{ 0.910,  0.953,  0.855}
\definecolor{green2}{rgb}{0.82,  0.902,  0.710}
\definecolor{green3}{rgb}{0.713,  0.903,  0.648}
\definecolor{green4}{rgb}{ 0.725,  0.855,  0.561}

\definecolor{defyellow}{rgb}{1,  0.983,  0.717}
\definecolor{defyellowtext}{rgb}{1,  0.851,  0.438}

\usepackage[colorlinks,linkcolor=black]{hyperref}
\title{HINT: Composed Image Retrieval with Dual-Path Compositional Contextualized Network}
\name{$^{1}$Mingyu Zhang
        \! $^{1}$Zixu Li
        \! $^{1}$Zhiwei Chen
        \!  $^{1}$Zhiheng Fu
        \! $^{1}$Xiaowei Zhu
         \! $^{1}$Jiajia Nie
        \! $^{1}$Yinwei Wei
        \! $^{1}$Yupeng Hu*\thanks{*Corresponding Author.}}

\address{$^{1}$ School of Software, Shandong University}

\begin{document}
%\begin{CJK}{UTF8}{gbsn}
\topmargin=0mm
%\ninept
%
\maketitle

\begin{abstract}
Composed Image Retrieval (CIR) is a challenging image retrieval paradigm. It aims to retrieve target images from large-scale image databases that are consistent with the modification semantics, based on a multimodal query composed of a reference image and modification text. Although existing methods have made significant progress in cross-modal alignment and feature fusion, a key flaw remains: \textbf{the neglect of contextual information in discriminating matching samples}. However, addressing this limitation is not an easy task due to two challenges: \textbf{1) implicit dependencies} and \textbf{2) the lack of a differential amplification mechanism}. To address these challenges, we propose a dual-pat\textbf{H} compos\textbf{I}tional co\textbf{N}textualized ne\textbf{T}work (\textbf{HINT}), which can perform contextualized encoding and amplify the similarity differences between matching and non-matching samples, thus improving the upper performance of CIR models in complex scenarios. Our HINT model achieves optimal performance on all metrics across two CIR benchmark datasets, demonstrating the superiority of our HINT model.
Codes are available at~\href{https://github.com/zh-mingyu/HINT}{https://github.com/zh-mingyu/HINT}.

\end{abstract}

\vspace{-0.5em}
\section{Introduction}

\begin{figure}[t!]
  \begin{center}
\includegraphics[width=0.78\linewidth]{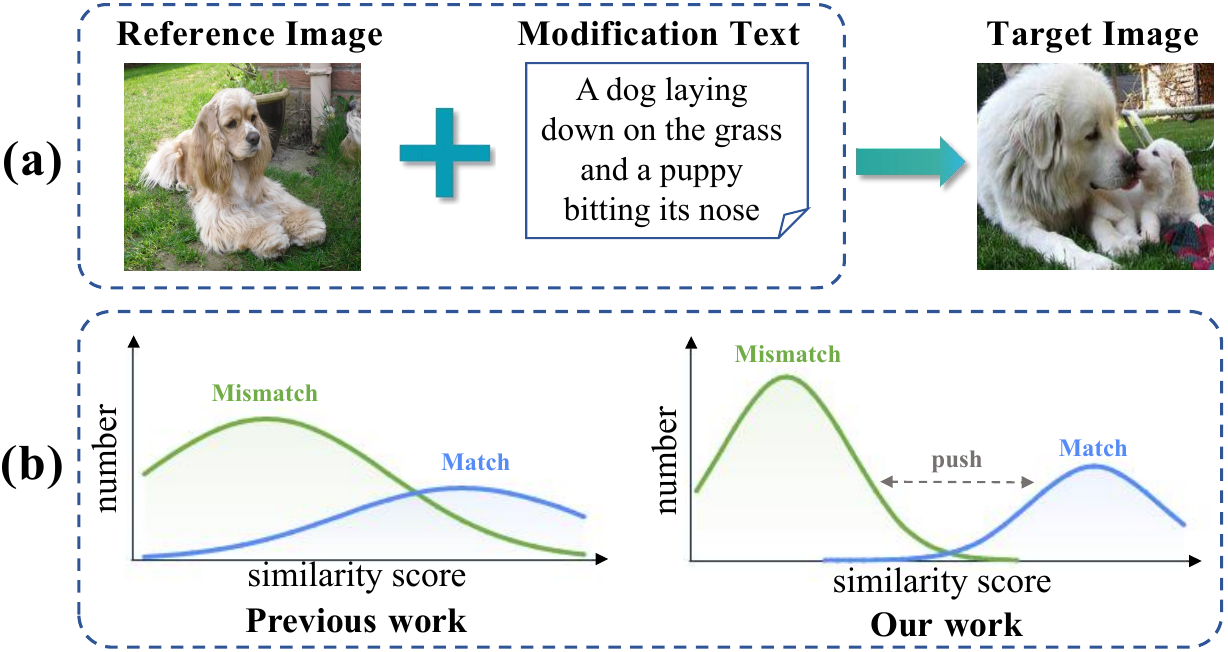}
  \end{center}
    \vspace{-20pt}
  \caption{An example of the CIR task and HINT's main idea.}
  \vspace{-10pt}
  \label{fig:intro}
  \vspace{-7pt}
\end{figure}

\noindent Composed Image Retrieval (CIR)~\cite{OFFSET,INTENT,FineCIR} aims to retrieve target images from large-scale image databases that are consistent with the modification semantics, based on a multimodal query composed of a reference image and modification text. Fig.~\ref{fig:intro}(a) is an example of this task. Unlike traditional retrieval, the text in CIR is typically a modification instruction for the reference image rather than a full description. This characteristic makes CIR more aligned with human interaction logic, thus holding significant application value in visual understanding~\cite{hu2021coarse,HABIT,xu2025hdnet,li2023ultrare}, video composition~\cite{REFINE,ReTrack,HUD}, and interactive search~\cite{hu2021video,hu2023semantic,tian2025core,tian2025open}.

Existing methods~\cite{tgcir,sprc,pair,median} have made significant progress in cross-modal alignment and feature fusion, but still suffer from a key limitation: the neglect of contextual information in discriminating matching samples. Specifically, existing methods~\cite{pair, median} mostly rely on global or local alignment to characterize the relationship between images and text, yet lack systematic modeling of contextual structure. The dependencies between regions within the reference image and the directive constraints of the modification text on local semantics are often simplified to shallow attention mechanisms. As shown in Fig.~\ref{fig:intro}(b), this causes the model to struggle in understanding complex modifications, weakening its ability to distinguish matching and non-matching samples, which may lead to misjudgments of non-matching images, directly limiting the upper performance of CIR models in complex scenarios.
However, addressing the aforementioned limitations is non-trivial due to the following two challenges. 
\textbf{1) Implicit dependencies}. Capturing the intra-modal context of the image faces challenges from background noise and regional heterogeneity, making it difficult to discern implicit dependencies across different local regions. Thus, how to effectively extract contextual semantics and quantify their relevance constitutes the first challenge. 
\textbf{2) Absence of the discrepancy-amplification mechanism}. In complex modification scenarios, models often exhibits blurred decision boundaries between matching and non-matching instances. This is because existing CIR methods typically use cosine similarity to assess the alignment between multimodal queries and target images, yet lack mechanisms to amplify the discrepancy between positive and negative pairs. Consequently, the second challenge lies in how to leverage contextual relevance to jointly optimize the training process, ensuring that the similarity of matched pairs is consistently enhanced while simultaneously suppressing that of mismatched pairs.

To address the above challenges, we propose a dual-pat\textbf{\underline{H}} compos\textbf{\underline{I}}tional co\textbf{\underline{N}}textualized ne\textbf{\underline{T}}work (\textbf{HINT}), which is capable of performing contextual encoding and amplifying the similarity differences between matching and non-matching samples, thereby improving the upper performance of CIR models in complex scenarios.
HINT consists of three modules. First, we design the \textit{Dual Context Extraction (DCE)} module, which extracts both intra-modal context and cross-modal context, enhancing joint semantic representation by integrating multimodal contextual information. 
Second, we introduce the \textit{Quantification of Contextual Relevance (QCR)} module, which measures the relevance between cross-modal contextual information and the target image semantics, enabling the quantification of the implicit dependencies. 
Finally, we develop the \textit{Dual-Path Consistency Constraints (DPCC)}, which optimizes the training process by constraining the representation consistency between multimodal fusion features and the target, ensuring the stable enhancement of the similarity for matching instances while lowering the similarity for non-matching instances. Notably, HINT achieves state-of-the-art performance on most metrics across two CIR benchmarks, FashionIQ and CIRR, demonstrating the superiority of our HINT model and the effectiveness of each module.

\begin{figure*}[ht]
\vspace{-10pt}
  \begin{center}
  \includegraphics[width=0.9\linewidth]{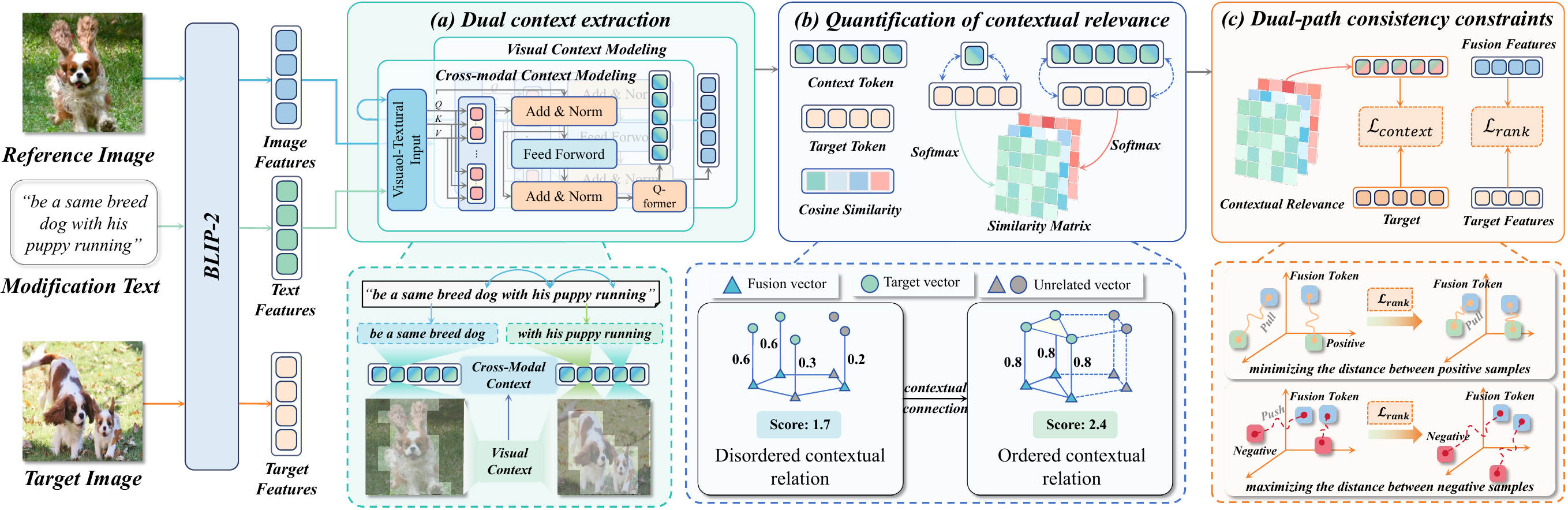}
  \end{center}
  \vspace{-20pt}
  \caption{\small HINT framework:~\textbf{(a)}~Dual~Context Extraction,~\textbf{(b)}~Quantification of Contextual Relevance,~\textbf{(c)} Dual-Path Consistency Constraints.}
  \vspace{-15pt}
  \label{fig:HINT}
\end{figure*}

\section{Methodology}

\noindent This paper proposes a dual-patH composItional coNtextualized neTwork (HINT), which is capable of extracting two types of contextual information from multimodal queries, including the visual intra-modal context for capturing the associations between regions within the image, and the cross-modal context for modeling the multimodal joint semantics between the image and text. 
The model then quantifies the relevance between the target image and the context based on contextual semantics and co-optimizes with the cosine similarity score, amplifying the similarity differences between matching and non-matching samples, thereby improving the upper performance of the CIR model in complex scenarios.
As shown in Fig.~\ref{fig:HINT}, HINT mainly consists of the following three modules: \textit{(a) Dual Context Extraction (DCE)}, which is used to extract dual contexts from multimodal queries, enhancing joint semantic representation; \textit{(b) Quantification of Contextual Relevance (QCR)}, which is used to evaluate the relevance between cross-modal context and the target image; \textit{(c) Dual-Path Consistency Constraints (DPCC)}, which leverages contextual relevance to collaboratively optimize the training process.

\subsection{Problem Formulation}
Given a collection of $N$ triplets, $\mathcal{T} = \left\{ \left( x_r, t_m, x_t \right)_n \right\}_{n=1}^{N}$, where $x_r$, $t_m$, and $x_t$ denote the reference image, modification text, and target image, respectively, the goal is to learn an embedding space in which the multimodal query $(x_r, t_m)$ is mapped as closely as possible to its corresponding target image $x_t$. The embedding function to be learned is formalized as $\mathcal{F}(x_r, t_m) \rightarrow \mathcal{F}(x_t)$.

\subsection{Dual Context Extraction (DCE)}
\noindent To address the implicit dependency, we conceptualize it as contextual semantics and introduce the \textit{Dual Context Extraction (DCE)}, which is capable of extracting context and enhancing cross-modal joint semantic representation through the fusion of contextual information.

\noindent\textbf{Visual Context Modeling (VCM).} 
Since images typically contain richer fine-grained information than text, 
and their local regions exhibit unordered structures with implicit inter-region dependencies, we first design VCM to model the visual intra-modal context, denoted as $\hat{\bm{V}}$.
Specifically, to leverage the prior knowledge of vision-language pre-trained models, we first use BLIP-2~\cite{blip-2} to extract high-quality features from the reference image, represented as $\bm{V} \in \mathbb{R}^{Q \times D}$, where $D$ denotes the feature dimension and $Q$ is the number of learnable queries. Similarly, we encode the modification text using BLIP-2 to obtain the text features $\bm{T}$.

In addition, to learn the local structural information within the image, we further perform region-level structured processing on the image. 
Specifically, we add position encodings to each of the $Q$ queries of the reference image $[\bm{v}_r^1,...,\bm{v}_r^Q]$,    formulated as,
\begin{equation}
\bm{V}_r^i =  \bm{v}_r^i  + \bm{P}_i,
\end{equation}
where $\bm{V}_r^i$ represents the feature of the $i$-th image query, and $\bm{P}_i$ denotes the position encoding for the $i$-th query. 
The final enhanced visual feature is constructed as $\bm{V} = [\bm{V}_{r}^1, ..., \bm{V}_{r}^Q] \in \mathbb{R}^{Q \times D}$.
Then, we perform attention-based interaction on the visual features to model the visual intra-modal context. The calculation process is represented as follows,
\begin{equation}
\fontsize{9pt}{9pt}
\bm{V}' = \text{LN}\left( \text{MSA}(\bm{V})+\bm{V} \right),
%\end{equation}
%\begin{equation}
\fontsize{9pt}{9pt}
\hat{\bm{V}} = \text{LN}\left( \bm{V}' + \text{FFN}\left( \bm{V}' \right)\right),
\end{equation}
where MSA denotes the multi-head attention layer, LN represents layer normalization, and the feed-forward network (FFN) consists of two linear layers and a ReLU function.

\noindent\textbf{Cross-modal Context Modeling (CCM).}
Given that text inherently possesses sequential order, it can be regarded as an ordered sequence of tokens, where the token arrangement naturally encodes contextual information. Therefore, we directly treat the text feature $\bm{T}$ as the textual intra-modal context.

To semantically fuse the visual intra-modal context $\hat{\bm{V}}$ and the textual intra-modal context $\bm{T}$ in order to obtain fine-grained cross-modal context features, we further introduce the Cross-modal Context Modeling (CCM).
Specifically, we first concatenate the visual intra-modal context $\hat{\bm{V}}$ with the text features $\bm{T}$ to form a multimodal joint representation $\bm{U} = [\hat{\bm{V}}, \bm{T}]$. Then, we process this joint representation using the same process as in VCM, as formulated below, 
\begin{equation}
\fontsize{9pt}{9pt}
\bm{U}' = \text{LN}\left(\text{MSA}(\bm{U}) + \bm{U}\right),
\bm{U}'' = \text{LN}\left(\bm{U}' + \text{FFN}(\bm{U}')\right),
\end{equation}
where $\bm{U}''$ represents the fine-grained cross-modal context, which encompasses the contextual information of both the reference image and the modification text. 
To further obtain context-aware multimodal fusion features, we leverage BLIP-2's Q-former architecture to learn joint semantic representation from the cross-modal context $\bm{U}''$, formulated as:
\begin{equation}
\fontsize{9pt}{9pt}
\hat{\bm{U}} = \text{Q-former}\left(\bm{U}''\right),
\end{equation}
where $\hat{\bm{U}} \in \mathbb{R}^{Q \times D}$ denotes the context-aware multimodal fusion features, and $Q$ is the number of Q-former's learnable queries.

\subsection{Quantification of contextual relevance (QCR)}
To evaluate the relevance between the aforementioned cross-modal contextual information and the semantics of the target image, and to enhance the model's ability to distinguish sample discrepancies, we compute a quantification score between the cross-modal context and the target image, which serves as the foundation for subsequent collaborative optimization.

Specifically, we first use BLIP-2 to extract the target image features, denoted as $\bm{F}$.
Then, for the context-aware multimodal fusion features $\hat{\bm{U}}$ extracted by the DCE module, we compute its incremental similarity score with the target image features $\bm{F}$ to quantify the contextual relevance. For $\hat{\bm{U}}$, we define its incremental similarity with the target image by calculating the similarity score between the first $k$ channels of the context-aware multimodal fusion features $\hat{\bm{U}}$ and the target features $\bm{F}$, as formulated below,
\begin{equation}
\fontsize{9pt}{9pt}
\bm{S}_k^u = \bm{S}(\hat{\bm{U}}, \bm{F})_k = \text{softmax}\left( \hat{\bm{U}}[:k] \cdot \bm{F}^T \right), k \in [1,...,Q],
\end{equation}
where $Q$ is the channel number of $\hat{\bm{U}}$, and $\bm{S}_k^u \in \mathbb{R}^{k \times Q}$ represents the $k$-th incremental similarity computation result. We perform the above operation $Q$ times, ultimately obtaining $[\bm{S}_1^u, ..., \bm{S}_Q^u]$.

Then, to align the similarity scores, we average the $\bm{S}_k^u$ obtained from the $k$-th computation along the channel dimension, resulting in the aligned incremental similarity score $\bar{\bm{S}}_k^u \in \mathbb{R}^{1 \times Q}$.
Finally, we average $[\bar{\bm{S}}_1^u, ..., \bar{\bm{S}}_Q^u]$ to fuse the multi-level contextual relevance quantification scores, as formulated below,
\begin{equation}
\fontsize{9pt}{9pt}
\bm{S}^u  = \text{mean}\left(\bar{\bm{S}}_1^u, ..., \bar{\bm{S}}_Q^u \right), 
\end{equation}
where $\bm{S}^u \in \mathbb{R}^{1 \times Q}$ represents the relevance quantification score between the cross-modal context and the target image, which will subsequently serve as a supplement to the original cosine similarity for co-optimization.

\subsection{Dual-path consistency constraints (DPCC)}
To leverage contextual relevance for co-optimizing the training process, ensuring the stable enhancement of similarity for matching instance pairs while lowering the similarity for non-matching instance pairs, we propose \textit{Dual-Path Consistency Constraints (DPCC)}.

Specifically, based on the relevance quantification score $\bm{S}^u$ obtained from the QCR module, we first design a Contextual Contrastive Loss, which is capable of optimizing the training process by constraining the representation consistency between multimodal feature contextual information and the target, as formulated below,
\begin{equation}
\fontsize{9pt}{9pt}
\mathcal{L}_{\text{context}} = \frac{1}{B} \sum_{i=1}^{B}- \log \frac{\exp \left( \bm{S}_i^u / \tau \right)}{\sum_{j=1}^{B} \exp \left( \bm{S}_i^u / \tau \right)},
\end{equation}
where, $B$ denotes the batch size, $\tau$ is the temperature coefficient, and $\bm{S}_i^u$ represents the relevance quantification score of the \$i\$-th triplet in the mini-batch. Additionally, we introduce Rank Loss~\cite{batch-based-classification-loss} to further enhance representation consistency, formulated as follows, 
\begin{equation}
\fontsize{9pt}{9pt}
\mathcal{L}_{rank} = \frac{1}{B} \sum_{i=1}^{B} -\log \left\{ \frac{\exp \left\{ \operatorname{s} \left( \overline{\textbf{U}}_{i} , \overline{\textbf{F}}_{i} \right)  / \tau\right\}}{ \sum_{j=1}^{B} \exp \left\{ \operatorname{s} \left( \overline{\textbf{U}}_{i}, \overline{\textbf{F}}_{j} \right) / \tau \right\}  } \right\},
\end{equation}
where $B$ denotes the batch size, $s(., .)$ represents the cosine similarity, and $\bar{\bm{U}}_i, \bar{\bm{F}}_i$ represent the average-pooled multimodal fusion features $\hat{\bm{U}}$ and the target image features $\bm{F}$ for the $i$-th triplet, respectively.
Finally, we define the overall optimization objective function as:
\begin{equation}
\fontsize{9pt}{9pt}
\mathcal{L} = \mathcal{L}_{\text{rank}} + \lambda \mathcal{L}_{\text{context}}.
\end{equation}
where $\lambda$ is the hyper-parameter that balances the two loss terms.
\section{Experiments}
\subsection{Experimental settings}\label{sec:experiment setting}

\noindent\textbf{Datasets}.
We evaluate the proposed HINT on an open-domain dataset CIRR~\cite{cirplant} and a fashion-domain dataset FashionIQ~\cite{FashionIQ}. Following previous work, we adopt the same evaluation metrics for both CIRR and FashionIQ: for CIRR, we evaluate R@$k$ ($k\!\!=\!\! 1, 5, 10, 50$), R$_{subset}$@$k$ ($k\!\!=\!\! 1, 2, 3$) and the average (mean of R@$5$ and R$_{subset}$@$1$). For FashionIQ, we evaluate R@$10$, R@$50$ for each category, as well as their averages.

\noindent\textbf{Implementation Details}. 
We use the pre-trained BLIP-2~\cite{blip-2} to extract features and fine-tune with an initial learning rate of $1e-6$. HINT is trained using the AdamW optimizer with an initial learning rate of $2e-5$, and the hidden dimension $D$ is set to $256$. The hyper-parameter $\lambda$ for the loss function is set to $0.2$. The model training is performed on a single NVIDIA V100 GPU with $32$GB of memory.

% \vspace{-0.5em}
\begin{table*}[ht]
  \centering
  \caption{Performance comparison on FashionIQ and CIRR relative to R@$k$(\%). The overall best results are in bold, while the second-best results are underlined. The Avg metric in CIRR denotes (R@$5$ + R$_{subset}$@$1$) / 2.} 
  \vspace{0pt}
        \resizebox{\linewidth}{!}{
        
    \begin{tabular}{l|cc|cc|cc|cc|cccc|ccc|c}
        \Xhline{2pt}
    \hline
    \multicolumn{1}{c|}{\multirow{3}{*}{Method}} &  \multicolumn{8}{c|}{FashionIQ}                              & \multicolumn{8}{c}{CIRR} \\
\cline{2-17}                  & \multicolumn{2}{c|}{Dresses} & \multicolumn{2}{c|}{Shirts} & \multicolumn{2}{c|}{Tops\&Tees} & \multicolumn{2}{c|}{Avg} & \multicolumn{4}{c|}{R@$k$} & \multicolumn{3}{c|}{R$_{subset}$@$k$} & \multirow{2}{*}{Avg} \\
\cline{2-16}                  & R@$10$  & R@$50$  & R@$10$  & R@$50$  & R@$10$  & R@$50$  & R@$10$ & R@$50$  & $k$=$1$   & $k$=$5$   & $k$=$10$ & $k$=$50$ & $k$=$1$   & $k$=$2$  & $k$=$3$ &  \\
    \hline
    \hline
    TG-CIR~\cite{tgcir}\textcolor{gray}{\scriptsize{(ACM MM'23)}}&  45.22 & 69.66 & 52.60 & 72.52 & 56.14 & 77.10 & 51.32 & 73.09 & 45.25 & 78.29 & 87.16 & 97.30 & 72.84 & 89.25 & 95.13 & 75.57 \\
    SSN~\cite{ssn}\textcolor{gray}{\scriptsize{(AAAI'24)}} &  34.36 & 60.78 & 38.13 & 61.83 & 44.26 & 69.05 & 38.92 & 63.89 & 43.91 & 77.25 & 86.48 & 97.45 & 71.76 & 88.63 & 95.54 & 74.51 \\
    
    MMF~\cite{mmf}\scriptsize{\textcolor{gray}{(ICASSP'24)}}&40.45  & 65.29  & 44.85  & 66.29  & 49.26  & 70.98  & 44.85  & 67.52 &43.80  & 77.37  & 86.94  & 97.71  & 71.39  & 88.16  & 95.15  & 74.38  \\
    CALA~\cite{cala}\textcolor{gray}{\scriptsize{(SIGIR'24)}}  & 42.38 & 66.08 & 46.76 & 68.16 & 50.93 & 73.42 & 46.69 & 69.22 & 49.11 & 81.21 & 89.59 & 98.00 & 76.27 & 91.04 & 96.46 & 78.74 \\
    SPRC~\cite{sprc}\textcolor{gray}{\scriptsize{(ICLR'24)}}  & \underline{49.18} & \underline{72.43} & \underline{55.64} & 73.89 & \underline{59.35} & 78.58 & \underline{54.72} & \underline{74.97} & 51.96 & 82.12 & 89.74 & 97.69 & \textbf{80.65} & \textbf{92.31} & \underline{96.60} & \underline{81.39} \\

    IUDC~\cite{iudc}\textcolor{gray}{\scriptsize{(TOIS'25)}} &35.22& 61.90 &41.86& 63.52& 42.19& 69.23& 39.76& 64.88 & -     & -     & -     & -     & -     & -     & -     & -\\
    COPE~\cite{cope}\textcolor{gray}{\scriptsize{(ACL'25)}} & 39.85 & 66.98 & 45.03 & 66.81 & 48.61 & 72.01 & 44.50 & 68.60 & 49.18 & 80.65 & 89.86 & 98.05 & 72.34 & 88.65 & 95.30 & 76.49 \\

    PAIR~\cite{pair}\scriptsize{\textcolor{gray}{(ICASSP'25)}} & 46.78  & 70.93  & 52.60  & 73.80  & 58.91  & \underline{78.81}  & 52.76  & 74.51 & 46.36 & 78.43 & 87.86 & 97.90 & 74.63 & 89.64 & 95.61 & 76.53\\
    QuRe~\cite{qure}\textcolor{gray}{\scriptsize{(ICML'25)}}& 46.80 & 69.81 & 53.53 & 72.87 & 57.47 & 77.77 & 52.60 & 73.48 & \underline{52.22} & \underline{82.53} & \underline{90.31} & \underline{98.17} & 78.51 & \underline{91.28} & 96.48 & 80.52 \\
    MEDIAN~\cite{median}\scriptsize{\textcolor{gray}{(ICASSP'25)}} & 46.90  & 70.30  & 52.65  & \underline{73.96}  & 57.62  & 78.63  & 52.39  & 74.30 & 45.66 & 78.72 & 87.88 & 97.89 & 75.52 & 89.45 & 95.57 & 77.12\\
    \hline
    \textbf{HINT (Ours)}  & \textbf{51.80} & \textbf{73.13} & \textbf{60.27} & \textbf{78.48} & \textbf{63.37} & \textbf{82.19} & \textbf{58.48} & \textbf{77.93} & \textbf{52.34} & \textbf{82.65} & \textbf{90.50} & \textbf{98.39} & \underline{80.48} & \textbf{92.31} & \textbf{96.77} & \textbf{81.57} \\
    \hline
        \Xhline{2pt}
    \end{tabular}%
    }
    \vspace{-10pt}
  \label{tab:main}%
\end{table*}%

\begin{table}[t!]
  \centering
  \caption{Ablation study on FashionIQ and CIRR datasets. We compute Avg-R@$10$, Avg-R@$50$ for FashionIQ, and Avg (mean of R@$5$ and R$_{subset}$@$1$) for CIRR, respectively.}
  % \vspace{-2pt}
  \resizebox{0.94\linewidth}{!}{
    \begin{tabular}{l|cc|cc|cc}
    \Xhline{2pt}
    \multicolumn{1}{c|}{\multirow{2}{*}{Method}} & \multicolumn{4}{c|}{FashionIQ} & \multicolumn{2}{c}{CIRR} \\
\cline{2-7}    
    \multicolumn{1}{c|}{} & R@$10$ & \multicolumn{1}{c|}{$\Delta$} & R@$50$ & \multicolumn{1}{c|}{$\Delta$} & Avg   & $\Delta$ \\
    \hline
    \hline
    \rowcolor[rgb]{ .949,  .949,  .949} \multicolumn{7}{c}{\textit{Dual Context Extraction (DCE)}}\\
    w/o VCM & 56.86  & \cellcolor{green1}-1.62  & 77.03  & \cellcolor{green2}-0.90  & 80.16  & \cellcolor{green2}-1.41 \\
    w/o CCM & 56.66  & \cellcolor{green2}-1.82  & 77.07  & \cellcolor{green1}-0.86  & 80.21  & \cellcolor{green1}-1.36 \\
    w/o DCE & 55.35  & \cellcolor{green3}-3.13  & 75.91  & \cellcolor{green3}-2.02  & 79.96  & \cellcolor{green3}-1.61 \\
    
   \rowcolor[rgb]{ .949,  .949,  .949} \multicolumn{7}{c}{\textit{Quantification of contextual relevance (QCR)}}\\
    w/o MC  & 56.48  & \cellcolor{green1}-2.00  & 76.92  & \cellcolor{green1}-1.01  & 80.50  & \cellcolor{green1}-1.07 \\
    w/o QCR & 55.16  & \cellcolor{green2}-3.32  & 76.54  & \cellcolor{green2}-1.39  & 80.07  & \cellcolor{green2}-1.50 \\
    
    \rowcolor[rgb]{ .949,  .949,  .949} \multicolumn{7}{c}{\textit{Dual-path consistency constraints (DPCC)}}\\
    w/o $\mathcal{L}_{rank}$    &  51.58  &  \cellcolor{green2} -6.90  &  72.62  & \cellcolor{green2} -5.31  &  77.15 & \cellcolor{green2} -4.42    \\
    w/o $\mathcal{L}_{context}$ &  53.54  &  \cellcolor{green1} -4.94  &  73.84  & \cellcolor{green1} -4.09  &  80.16 & \cellcolor{green1} -1.41    \\
    \hline
    \hline
    \multicolumn{1}{l|}{\textbf{HINT~(Ours)}} 
    &  \textbf{58.48}
    & \multicolumn{1}{c|}{\cellcolor[rgb]{ 1,  0.983,  0.717}   -0.00} 
    &  \textbf{77.93 }     
    & \multicolumn{1}{c|}{\cellcolor[rgb]{ 1,  0.983,  0.717}   -0.00}          
    &  \textbf{81.57 }      
    & \cellcolor[rgb]{ 1,  0.983,  0.717}   -0.00         \\
    \Xhline{2pt}
    \end{tabular}%
  }
  \vspace{-10pt}
  \label{tab:abla}%
\end{table}%

\subsection{Perfermance Comparison}
We systematically present the performance analysis results of the proposed HINT in Table~\ref{tab:main}. Through experiments on two different types of datasets, we draw the following key conclusions:
1) HINT demonstrates significant performance improvement on both datasets, validating its strong cross-domain generalization ability. Specifically, compared to suboptimal models, HINT achieves an improvement in R@$1$ on CIRR dataset, showcasing its outstanding performance in open-domain. 
Additionally, on FashionIQ dataset, HINT achieves a notable relative increase in the average R@$10$. 
This indicates that HINT can adapt stably and efficiently to different retrieval scenarios, whether in open-domain or fashion-domain.
2) The performance improvement of HINT on FashionIQ dataset is more pronounced than on CIRR dataset. This is because the CIR task in fashion-domain typically involves fine-grained and highly structured attribute composition, leading to more frequent semantic ambiguity and conflicts. 
By introducing a context-aware mechanism, HINT effectively alleviates such semantic ambiguities and better understands and integrates the deep semantic relationships between images and text.

\subsection{Ablation Study}

\noindent To evaluate the contribution of each component in our model, we conduct a comprehensive ablation study on the modules within HINT. 
Specifically, the ablation variants we design are as follows:
\begin{itemize}[leftmargin=*]
\item \textbf{w/o VCM}: In this variant, we remove the Visual Context Modeling (VCM) from HINT to assess the role of visual internal context.
\item \textbf{w/o CCM}: In this setup, we do not introduce Cross-modal Context Modeling (CCM) and compare the contribution of cross-modal context in the model.
\item \textbf{w/o DCE}: To verify the effectiveness of the DCE module, we remove it from the HINT framework and examine its impact on overall performance.
\item \textbf{w/o MC}: In the incremental similarity computation, we simplify the multi-channel (MC) multimodal fused features to a single channel and calculate the similarity score with the target features.
\item \textbf{w/o QCR}: We replace the original method for calculating association quantization scores with basic cosine similarity to evaluate its impact on visual retrieval results.
\item \textbf{w/o $\mathcal{L}_{rank}$} and \textbf{w/o $\mathcal{L}_{context}$}: To further investigate the contribution of different loss terms, we separately remove the batch-based contrastive loss $\mathcal{L}_{rank}$ and the context loss $\mathcal{L}_{context}$ to observe their independent effects during metric learning.
\end{itemize}

\noindent As shown in Table~\ref{tab:abla}, we observe the following results:
1) \textbf{w/o VCM}, \textbf{w/o CCM}, and \textbf{w/o DCE} all lead to a decrease in model performance. This is primarily due to the absence of contextual relationships, which impacts the model's semantic understanding and matching of images and text.
2) The performance of \textbf{w/o MC} and \textbf{w/o QCR} does not reach the optimal level, which validates the critical role of the association quantization score calculation method in the HINT model. The inter-modal context is crucial to the model's performance. Without context-aware similarity computation, the semantic relationship between images and text becomes unbalanced, which in turn affects the final retrieval results.
3) By comparing the experimental results of \textbf{w/o $\mathcal{L}_{rank}$} and \textbf{w/o $\mathcal{L}_{context}$}, we observe that the performance of \textbf{w/o $\mathcal{L}_{rank}$} significantly declines, highlighting the powerful ability of batch-based contrastive loss in the CIR task. The results of \textbf{w/o $\mathcal{L}_{context}$} indicate that without considering the contextual information, the performance of HINT deteriorates. This reflects the importance of contextual information in similarity computation, especially in multimodal tasks, where the lack of context optimization leads to excessive interference from local information, thus affecting the stability of the model.

\subsection{Case Study}

\begin{figure}[ht]
  \begin{center}
  \vspace{-5pt}
\includegraphics[width=0.85\linewidth]{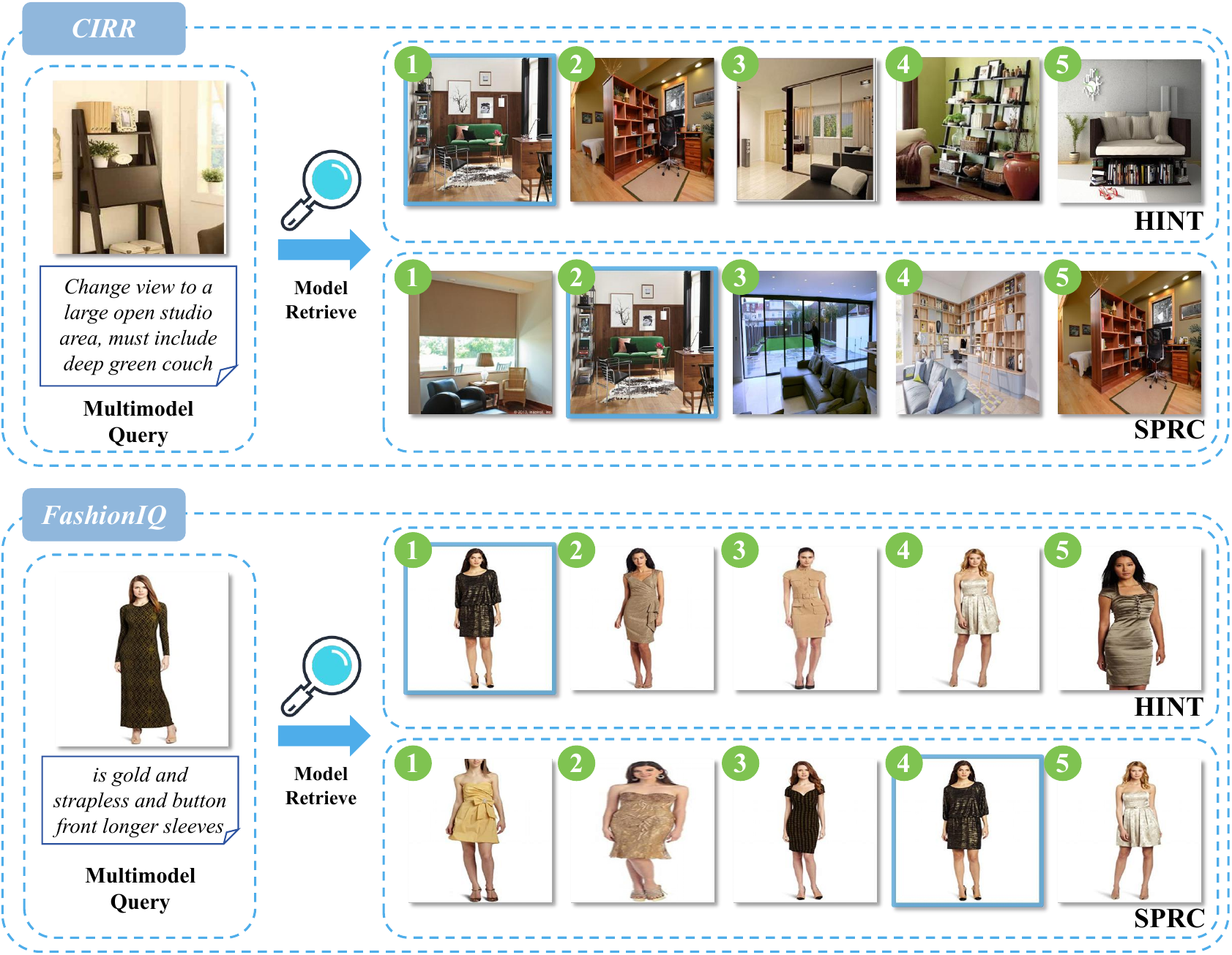}
  \end{center}
  \vspace{-18pt}
  \caption{\small Case study on CIRR and FashionIQ.}
  \vspace{-15pt}
  \label{fig:case}
\end{figure}

\noindent To validate the advantages of our HINT in terms of contextual information, we present the top five retrieval results of HINT and SPRC on the CIRR and FashionIQ datasets in Fig.~\ref{fig:case}, with the target image highlighted by a blue rectangle. The experimental results show that, on the CIRR dataset, HINT successfully retrieves the target image as the top result, while the latter ranks second. This indicates that HINT is better at capturing contextual information, thereby significantly improving retrieval accuracy. On the FashionIQ dataset, HINT also demonstrates exceptional performance. In contrast, SPRC fails to successfully retrieve the target image, possibly due to its inability to effectively handle the ambiguity in the semantic modifications of the text. HINT, on the other hand, shows superior capability in dealing with fine-grained attribute combinations and complex semantic relationships, especially in the fashion domain, where this advantage is particularly evident. These results further confirm the strong ability of the HINT model in handling semantic ambiguity or conflict, as well as fine-grained attribute combination tasks, reflecting its superiority in complex scenarios.

\section{Conclusion}

\noindent In this work, we focus on a key limitation of existing CIR models, which is the neglect of the role of contextual information in discriminating matching samples. To address this limitation, we further break it down into two key challenges: the lack of implicit dependencies and the absence of a differential amplification mechanism. Based on this, we propose a novel dual-patH composItional coNtextualized neTwork (HINT) to tackle the CIR task. HINT is capable of performing context-aware encoding and amplifying the similarity differences between matching and non-matching samples, thereby improving the upper performance of CIR models in complex scenarios. It achieves optimal results across all metrics on two CIR benchmark datasets.

\section{Acknowledgment}
This work was supported in part by the National Natural Science Foundation of China, No.:62276155, No.:62576195, and No.:62572282; in part by the China National University Student Innovation \& Entrepreneurship Development Program, No.:2025282 and No.:2025283.

\bibliographystyle{IEEEbib}
\bibliography{strings}
%\end{CJK}
\end{document}